\documentclass[sn-basic,iicol]{sn-jnl}% Default with double column layout

\usepackage{arydshln}
\usepackage{graphicx}%
\usepackage{multirow}%
\usepackage{amsmath,amssymb,amsfonts}%
\usepackage{amsthm}%
\usepackage{mathrsfs}%
\usepackage[title]{appendix}%
\usepackage{xcolor}%
\usepackage{textcomp}%
\usepackage{manyfoot}%
\usepackage{bbding}
\usepackage{booktabs}%
\usepackage{subfigure} 
\usepackage{algorithm}%
\usepackage{algorithmicx}%
\usepackage{algpseudocode}%
\usepackage{listings}%
\theoremstyle{thmstyleone}%
\begin{document}

\title{Efficient Vision-and-Language Pre-training with Text-Relevant Image Patch Selection}


\author{\fnm{Wei} \sur{Ye}$^\text{1}$}\email{wye@pku.edu.cn}
\equalcont{These authors contributed equally to this work.}

\author{\fnm{Chaoya} \sur{Jiang}$^\text{1}$}\email{jiangchaoya@pku.edu.cn}
\equalcont{These authors contributed equally to this work.}
\author*{\fnm{Haiyang} \sur{Xu}$^{\text{2}*}$}\email{shuofeng.xhy@alibaba-inc.com}
\author{\fnm{Qinghao} \sur{Ye}$^\text{2}$}\email{yeqinghao.yqh@alibaba-inc.com}
\author{\fnm{Chenliang} \sur{Li}$^\text{2}$}\email{lcl193798@alibaba-inc.com}
\author{\fnm{Ming} \sur{Yan}$^\text{2}$}\email{ym119608@alibaba-inc.com}
\author*{\fnm{Shikun} \sur{Zhang}$^{\text{1}*}$}\email{zhangsk@pku.edu.cn}
\author{\fnm{Songfang} \sur{Huang}$^\text{2}$}\email{songfang.hsf@alibaba-inc.com}
\author{\fnm{Fei} \sur{Huang}$^\text{2}$}\email{fei.huang@alibaba-inc.com}

\affil*[1]{\orgdiv{National Engineering Research Center for Software Engineering}, \orgname{Peking University}, \orgaddress{ \city{Beijing}, \country{China}}}

\affil[2]{\orgdiv{DAMO Academy}, \orgname{Alibaba Group}, \orgaddress{ \city{Beijing}, \country{China}}}


\abstract{Vision Transformers (ViTs) have become increasingly popular in large-scale Vision and Language Pre-training (VLP) models. Although previous VLP research has demonstrated the efficacy of ViTs, these efforts still struggle with computational inefficiencies caused by lengthy visual sequences. To address this challenge, we introduce an efficient VLP approach called TRIPS, which stands for \textbf{T}ext-\textbf{R}elevant \textbf{I}mage \textbf{P}atch \textbf{S}election. TRIPS progressively reduces the visual sequence using a text-guided patch-selection layer in the visual backbone, thereby accelerating both training and inference processes. This patch-selection layer dynamically computes text-dependent visual attention, enabling it to identify attentive image tokens with text guidance and fuse inattentive ones in an end-to-end fashion. Importantly, TRIPS does not add any extra parameters and generalizes to most ViT-based VLP models. We incorporate TRIPS into three representative VLP models covering single-stream, dual-stream, and generative paradigms, and conduct extensive experiments on five  widely-used multi-modal benchmark datasets. Our experimental results reveal that TRIPS delivers a 40\% speedup, while maintaining competitive or superior performance on downstream tasks.}




\maketitle

\section{Introduction}
\begin{figure*}[!t]
\centering

\centering
\includegraphics[width=0.98\linewidth]{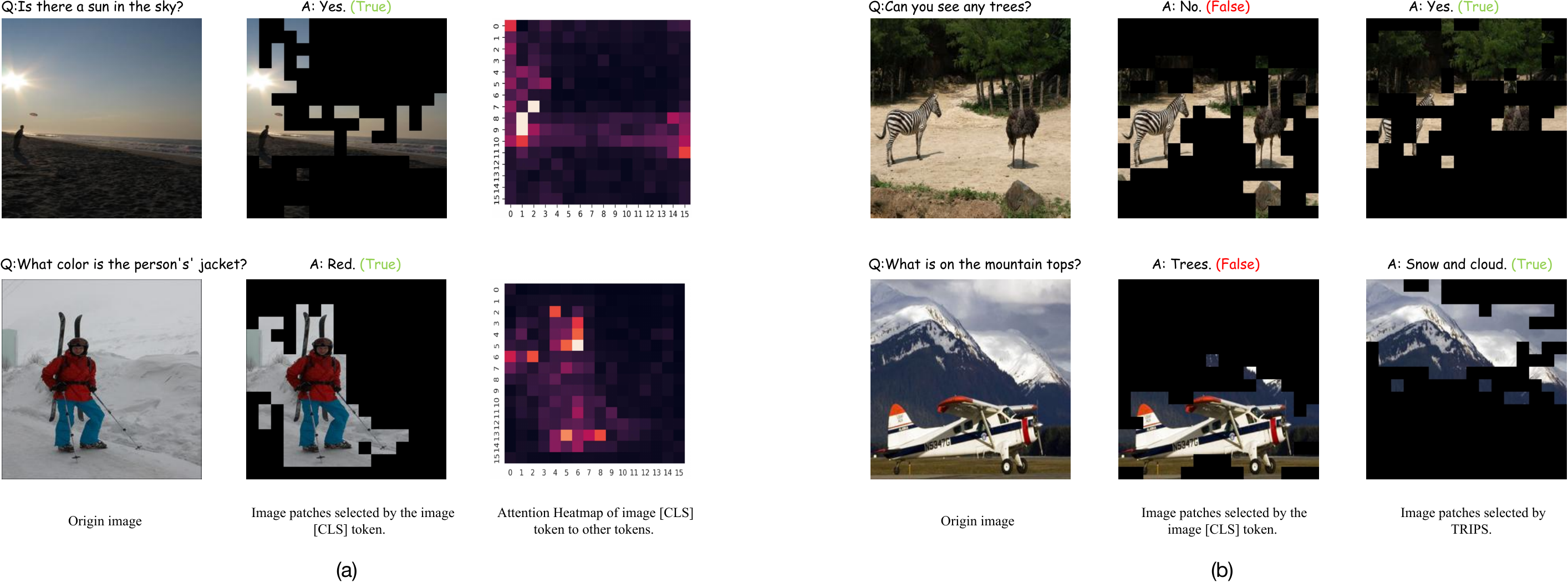}
\caption{
% The sub-figure (a) shows the VQA cases of the ALBEF~\citep{Li2021AlignBF} finetuned on the VQA task, in which the input image tokens are directly selected with the guidance of attention weight of the image [CLS] token to other image tokens. We visualize the attention distribution of the image [CLS] token. As we can see, the [CLS] token attention naturally focuses on the objects in images and ignores the image backgrounds. If the question is about the objects in the images, we can still get the correct prediction. The sub-figure (b) shows the VQA predictions comparison between ALBEF, which directly selects image tokens with the guidance of the image [CLS] token, and our model TRIPS. As we can see, the questions are about the image backgrounds, and the former predicts the wrong answers, yet the latter can give the correct answer as it preserves the text-relevant image tokens.
Sub-figure (a) presents the VQA cases for ALBEF~\citep{Li2021AlignBF} fine-tuned on the VQA task, where input image tokens are directly selected based on the attention weight of the image [CLS] token in relation to other image tokens. We visualize the attention distribution of the image [CLS] token, which, as seen, naturally concentrates on objects within the images while disregarding the backgrounds. If the question pertains to the objects in the images, accurate predictions can be obtained. Sub-figure (b) compares VQA predictions between ALBEF, which directly selects image tokens guided by the image [CLS] token, and our model, TRIPS-ALBEF. As illustrated, when questions relate to image backgrounds, the former produces incorrect answers, whereas the latter provides the right responses by preserving text-relevant image tokens.}
% \caption{ROUGE-1, ROUGE-2, and ROUGE-L Scores of simple summaries. Simple summaries are generated by straightforwardly concatenating aligned sentences (red) or golden captions (blue) of a document's first $k$ images. The scores are calculated by matching them against the reference summaries. The horizontal green dashed lines represent the text summaries generated by our SITA model.}
\label{fig1}
\end{figure*}
% Recent years, with the success of Transformer in computer vision field \citep{}, Transformer-based model   are served as the vision backbone in more and more Vision-Language(VL) pre-training research \citep{}. However, 
In recent years, Vision-Language Pre-training (VLP)  \citep{Tan2019LXMERTLC,Chen2019UNITERLU,Lu2019ViLBERTPT,Huang2020PixelBERTAI,Li2020OscarOA,Chen2020UNITERUI,Zhou2020UnifiedVP,Li2021AlignBF,Yu2021ERNIEViLKE} has experienced remarkable growth, emerging as a dominant paradigm for addressing Vision-Language (VL) tasks. Conventional VLP models \citep{Tan2019LXMERTLC,Chen2019UNITERLU,Lu2019ViLBERTPT,Li2020OscarOA} employ pre-trained object detectors \citep{Ren2015FasterRT,Redmon2016YouOL,He2017MaskR} to extract region-based image features but suffer from extensive annotation and expensive computation of object detector training.
Inspired by the success of the Vision Transformer (ViT)~\citep{Dosovitskiy2021AnII} and its variants~\citep{Liu2021SwinTH,Wu2021CvTIC,Wang2021PyramidVT} in the computer vision domain, more recent VLP models ~\citep{Li2021AlignBF,Radford2021LearningTV,Kim2021ViLTVT,Wang2021VLMoUV,Singh2021FLAVAAF} have adopted ViT as the visual encoder or cross-modal fusion encoder without using region features from the pre-trained object detector.

However, ViT-based VLP methods necessitate the modeling of lengthy visual sequences, e.g., from high-resolution images, to achieve robust vision understanding, which results in quadratic computational complexity proportional to the visual sequence length. Furthermore, recent studies \citep{Dosovitskiy2021AnII, Touvron2021TrainingDI} have started to investigate vision-language foundation models, focusing on scaling up both model and data size. Consequently, there is an increasing need to reduce the substantial computational costs associated with ViT-based VLP models. As demonstrated in Figure \ref{fig1} (a), we observe that eliminating the inattentive patch tokens of the image [CLS] token generally does not impact the Visual Question Answering (VQA)~\citep{Agrawal2015VQAVQ} predictions. Based on this observation, we hypothesize that not all image tokens in the visual encoder positively contribute to the final prediction results of VLP models, and a considerable number of redundant image tokens exist.

Recent studies ~\citep{Rao2021DynamicViTEV,Xu2021EvoViTST,Zong2021SelfslimmedVT,Liang2022NotAP} have explored ViT model acceleration by eliminating unrelated image tokens. However, these methods are specifically tailored for computer vision tasks (e.g., image recognition) and remove redundant tokens based on visual semantics while disregarding the aligned knowledge in the text modality. Consequently, they are not suitable for VL tasks. As illustrated in Figure \ref{fig1} (b), removing patch tokens based solely on the image [CLS] token, without the guidance of text knowledge, leads to incorrect answers. This observation inspires us to reduce the number of image tokens by merging less informative patch tokens under the guidance of the aligned text context.

In this work, we introduce an efficient VLP model featuring Text-Relevant Image Patch Selection (TRIPS) to progressively reduce redundant image tokens with the guidance of text. TRIPS selects text-consistent image tokens through a text-aware patch-selection layer, thereby minimizing the computational cost of visual encoding and cross-modal fusion. This patch-selection layer preserves attentive image tokens with text guidance and fuses inattentive tokens into a single one by dynamically computing text-dependent visual attention in an end-to-end manner. Consequently, we gradually decrease the number of image tokens as the visual backbone deepens, alleviating the computational burden of the visual encoder and enhancing the efficiency of cross-modal fusion due to the reduced visual sequences. Additionally, model efficiency can be flexibly controlled by adjusting the keep rate of image tokens in the patch-selection layer without introducing any additional parameters. 

We evaluate TRIPS on five representative VL tasks consisting of visual question answering (VQA), natural language visual reasoning (NLVR), cross-modal retrieval, image captioning, and visual grounding. By incorporating TRIPS into existing representative VLP models, we achieve approximately 40\% efficiency gains while maintaining competitive or superior downstream task performance. For example, equipped with TRIPS, ALBEF~\citep{Li2021AlignBF}, a recent robust two-stream VLP model, can accelerate by 40.98\% and even improve by 0.1 on the VQA test-dev and 0.2 on the NLVR Dev (see Table \ref{table:vqa_caption}). Moreover, by increasing the input image resolution while maintaining the same computational cost, TRIPS can enhance performance by 0.4 on the VQA test-dev and 0.6 on the NLVR Dev.

Our main contributions can be summarized as three-fold:

\begin{itemize}

\item We introduce an efficient vision-and-language pre-training model featuring Text-Relevant Image Patch Selection (TRIPS). To the best of our knowledge, this is the first attempt to reduce the computational cost of VLP models by diminishing image tokens with the assistance of linguistic context.
\item We propose a novel text-relevant patch-selection layer, which can dynamically compute text-dependent visual attention to identify attentive (or critical) image tokens and merge inattentive ones with text guidance in an end-to-end manner.
\item Comprehensive experiments demonstrate that our TRIPS can enhance the training and inference efficiency of VLP models while incurring lower computational costs. Moreover, by increasing the input image resolution, TRIPS leverages additional image tokens to achieve superior performance without increasing computational expenses.
\end{itemize}

\section{Related Work} 
\subsection{Vision-Language Pre-training} 
Current approaches to VLP can be broadly divided into two categories in terms of visual representation extraction. The first category is detector-based VLP methods \citep{Lu2019ViLBERTPT,Li2019VisualBERTAS,Tan2019LXMERTLC,Li2020OscarOA,Chen2020UNITERUI,Yu2021ERNIEViLKE}. These methods primarily adopt a two-step training pipeline: they extract visual features using a pre-trained object detector and then train the cross-modal pre-training model to align text and visual features. Some region-based methods aim to reduce computational costs with lightweight model architectures \citep{wang2020minivlm,gan2021playing} or knowledge distillation \citep{fang2021compressing}. However, these methods still face several challenges, including expensive computation and time consumption for object/region detection and error propagation problems caused by the two-step pre-training strategy. The main challenge for these methods is to balance effectiveness and efficiency. The second category consists of more recent CNN-based \citep{Huang2020PixelBERTAI,Xu2021E2EVLPEV} or ViTs-based \citep{Li2021AlignBF,Kim2021ViLTVT,Radford2021LearningTV,Wang2021VLMoUV} methods, especially patch-based ViT. These methods eliminate the need for a complex object detector in feature extraction, enabling end-to-end VL learning. However, few works have focused on reducing the high computational cost of ViT-based VLP models.

In terms of fusion schemes for modeling cross-modal interaction, typical VLP approaches \citep{Tan2019LXMERTLC,Li2021AlignBF,Li2020OscarOA,Wang2021SimVLMSV,Kim2021ViLTVT,li2022blip,Wang2021VLMoUV} can also be categorized into \textit{dual stream} and \textit{single stream}. Single-stream approaches \citep{Li2020OscarOA,Kim2021ViLTVT,Wang2021SimVLMSV,Wang2021VLMoUV} assume that the potential correlation and alignment between the two modalities are relatively straightforward and can be learned by a single transformer encoder. As a result, the architecture concatenates text embeddings and image features, along with special embeddings that indicate their respective positions and modalities. This concatenated feature set is then fed into a transformer-based encoder for further processing. Dual-stream approaches \citep{Tan2019LXMERTLC,li2022mplug,li2022blip,dou2021empirical,Xu2021E2EVLPEV} assume that the intra-modal interaction and cross-modal interaction need to be separated to obtain better multimodal representations. Thus, they employ two single-modal encoders to separately encode images and text. Additionally, previous VLP methods \citep{Li2019VisualBERTAS, Chen2019UNITERLU, Su2020VLBERTPO, Wang2021VLMoUV} were typically only capable of performing tasks related to vision-language understanding and reasoning (e.g., VQA and NLVR). However, recent multimodal generation models \citep{li2022blip,li2022mplug,Wang2021SimVLMSV} have started to adopt encoder-decoder style generative approaches for accomplishing tasks associated with multimodal text generation.
% In this paper, we propose the text-relevant image patch selection mechanism to decrease the computation cost of VLP models by reducing the visual sequence progressively with a text-guided patch-selection layer in the visual backbone for efficient training and inference. 
Our method can easily generalize to most ViT-based VLP models, including single-stream, dual-stream, and generative approaches.

\subsection{ViTs Acceleration} 
Numerous studies have focused on proposing more efficient attention mechanisms \citep{Wang2020LinformerSW, Kitaev2020ReformerTE, Choromanski2021RethinkingAW} or compressing Transformer structures \citep{Liu2021SwinTH, Heo2021RethinkingSD,Wang2021PyramidVT} to accelerate the computation of transformer-based models \citep{Vaswani2017AttentionIA}. Recently, some approaches have aimed to accelerate ViTs by reducing the number of tokens involved in ViT inference. For instance, \citep{Ryoo2021TokenLearnerAS} proposed TokenLearner to expedite ViTs, in which a relatively small number of tokens are learned by aggregating the entire feature map weighted by dynamic attention. \citep{Rao2021DynamicViTEV} introduces a method to reduce tokens for a fully trained ViT, where an extra learnable neural network is added to ViT to select a subset of tokens. \citep{Liang2022NotAP} proposes to reduce the computational overhead of inference by introducing a token reorganization method to progressively reduce and reorganize image tokens. However, these methods are not suitable for VLP, as they reduce image tokens without considering the text context. In contrast, our proposed approach takes into account the textual context, making it more appropriate for VLP tasks.

\begin{figure*}[htpb]
\setlength{\belowcaptionskip}{-0.15cm}
\centering
\includegraphics[width=0.97\textwidth]{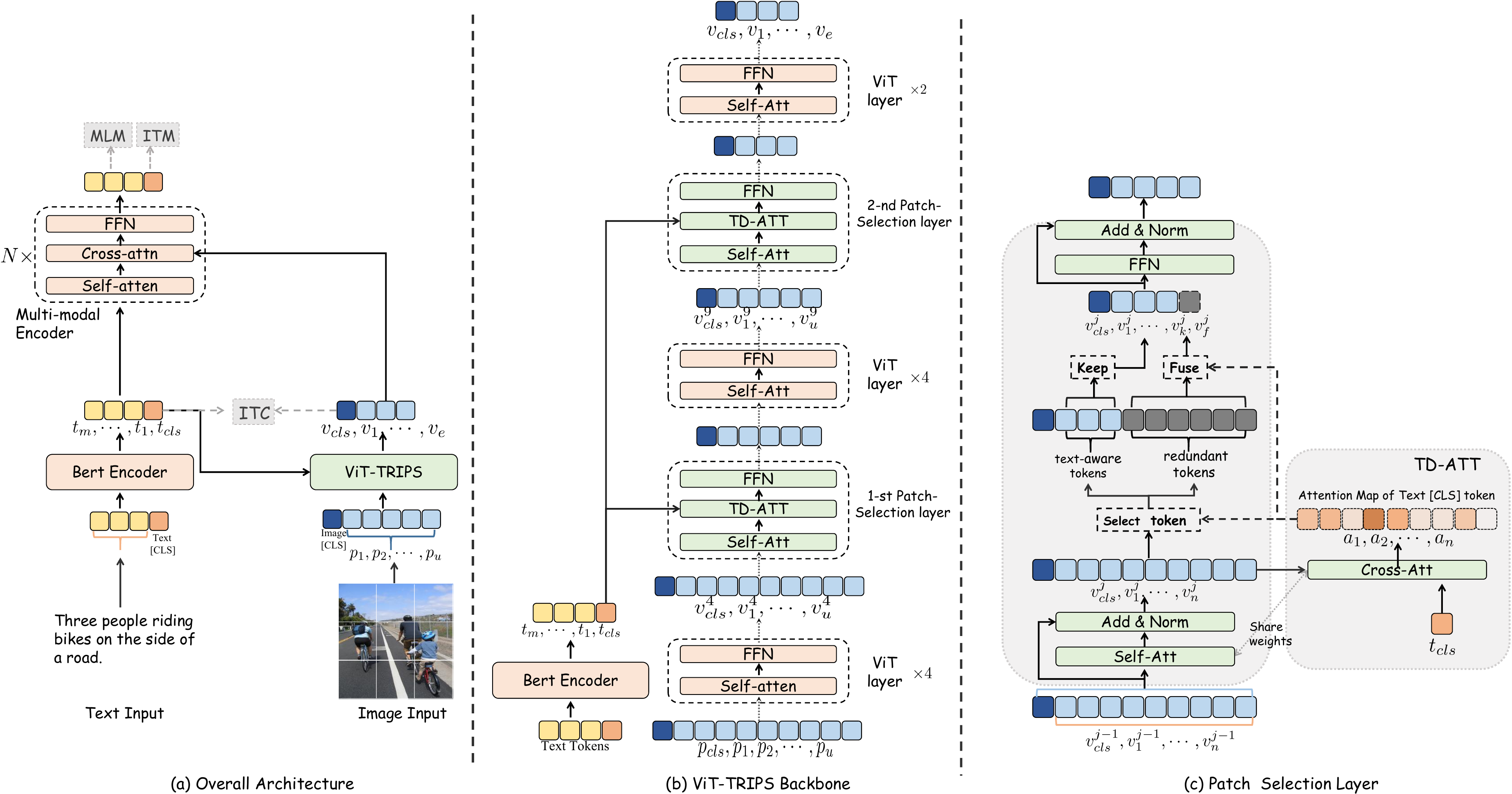}%fig2文件夹下的xbee.esp图片，

\caption{Sub-figure (a) showcases the overall architecture of the VLP model (TRIPS-ALBEF) presented in this paper. Sub-figure (b) provides a visual overview of the ViT augmented with a Text-Relevant Image Patch Selection module (ViT-TRIPS). We assume the ViT-TRIPS comprises 12 layers, and the 5th and 10th layers serve as the patch selection layers. Sub-figure (c) depicts the design of the Text-Relevant Image Patch Selection layer.}

\label{fig2}
\end{figure*}

\section{Method}

Our method aims to generalize to most ViT-based VLP models covering single-stream, dual-stream, and generative VLP paradigms. In this section, we mainly use the dual-stream ALBEF~\citep{Li2021AlignBF} framework as a representative base model to introduce the technical design of our TRIPS (named TRIPS-ALBEF). We will first present the model architecture, and then introduce the ViT-based visual backbone with the acceleration module of the text-relevant patch-selection layer. We will also briefly discuss how we incorporate TRIPS into a recent generative VLP model mPLUG~\citep{li2022mplug} and single-stream model ViLT~\citep{Kim2021ViLTVT}. Finally, we will introduce the pre-training objectives.

\subsection{Model Architecture}

% We focus on the acceleration for ViT-based VLP models and propose a plug-and-play module PASE which can be equipped with either single stream VLP model or dual stream VLP model. 
% The main current VLP models can be divided into three types: single-stream models, dual-stream models with cross-attention fusion, and dual-stream models with connect-attention fusion. 
%As shown in fig ~\ref{fig2}, we will first introduce the single-stream models with patch selection. As the difference between the dual-stream models with cross-attention fusion, and dual-stream models with concatenating fusion is in the decode stage, we introduce them in the same subsection. 
As shown in Figure~\ref{fig2} (a), TRIPS-ALBEF contains a visual encoder with the text-relevant patch-selection layer, a text encoder, and a multimodal fusion encoder. The visual encoder takes a  Vision Transformer (ViT), where text-relevant patch-selection layers are used to progressively reduce and reorganize image tokens, namely ViT-TRIPS. The text encoder adopts BERT$_{base}$ transformer \citep{Devlin2019BERTPO}. Similar to~\cite{Li2021AlignBF}, the multimodal fusion encoder is a transformer encoder that performs the cross-modal interaction and fusion through a cross-attention mechanism. 

Formally, given an input image-text pair, we first feed the input text to the text encoder and represent it as a sequence of embeddings $T = \{t_{cls}, t_1, t_2, \cdots, t_m \}$, where $t_{cls}$ is the embedding of the text [CLS] token to summarize the input text. Then, we divide the input image into patches $P = \{p_{cls}, p_1,p_2, \cdots, p_u \}$, and encode them with the image encoder ViT-TRIPS. It takes the text [CLS] embedding $t_{cls}$ and image patches $\{p_{cls}, p_1,p_2, \cdots, p_u \}$ as input, and outputs the image sequence $V=\{v_{cls}, v_1, v_2, \cdots,v_e\}$. Note that $e < u$, since we apply the text-relevant patch-selection layer to select the text-aware image tokens and fuse the redundant tokens, allowing us to reduce the total visual sequence length for efficiency. Finally, the text features $ \{t_{cls}, t_1, t_2, \cdots, t_m \}$  and the image features  $\{v_{cls}, v_1,v_2, \cdots, v_e \}$ encoded by the image encoder are fused by cross attention at each layer of the multimodal encoder as in ALBEF~\citep{Li2021AlignBF}. The output of the multimodal encoder is used to pre-train and finetune downstream tasks.

\noindent \subsection{Text-Relevant Image Patch Selection}\label{sect2}
Existing works in computer vision~\citep{Rao2021DynamicViTEV, Liang2022NotAP} select patches by using only the image [CLS] token from the ViT backbone. However, as shown in Figure \ref{fig1} (b), the selection of image tokens in cross-modal tasks is closely related to textual context, and different texts for a single image may focus on different parts of the visual content. Selecting the image tokens with the guidance of aligned textual content can help the VLP model focus on the key parts of the image for more effective and efficient cross-modal fusion. Here, we present a text-relevant patch-selection layer that can dynamically select the image patches with the guidance of textual input, yet with no additional parameters introduced.

As shown in Figure \ref{fig2} (b), for a ViT with $L$ standard Transformer layers and $t$ patch-selection layers in total, the interval length is obtained as $s = L / (t + 1)$. Then, we choose the layer index $j=i*s+1$ as the $i_{th}$ patch-selection layer, so that patch-selection layers are uniformly inserted into the ViT-TRIPS backbone. In each patch-selection layer, as shown in Figure \ref{fig2}(c), we adopt standard self-attention (SA), Text-aware Dynamic Attention (TD-ATT), and Inattentive Token Fusion (ITF) modules to progressively reduce image tokens.

Specifically, for the $i_{th}$ patch-selection layer, image features $v^{j-1}$ =
$\{v^{j-1}_{cls}, $ $v^{j-1}_1, \cdots, v^{j-1}_n\}$ are first fed to the $j_{th}$ visual self-attention layer:

\begin{equation}
v^{j}=LN(SA(v^{j-1})+v^{j-1})
\end{equation}
where LN is short for layer normalization, and $n$ is the number of patch tokens in the $j-1$ visual transformer layer. 
 Next, we will illustrate how to use the Text-aware Dynamic Attention mechanism (TD-ATT) to select the text-aware image patch tokens. The text [CLS] embedding $t_{cls}$ is linearly projected to the query vector denoted as $q_{text}$ by the shared query linear layer of the $j_{th}$ visual self-attention layer. We compute the text-to-image attention feature map excluding the image [CLS] token as follows:
% \begin{equation}
%     q_{text} =  \mathcal{Q}(t_{cls})
% \end{equation}
\begin{equation}
    a_{cls} = softmax(\frac{ q_{text} \cdot v^{j}[1:]^T}{\sqrt{d}})
\end{equation}

We identify and preserve the attentive image tokens
corresponding to the $k$ largest elements in the attention map $a_{cls}=\{a_{1}, ..a_{n} \}$, where $k = n \times r$, and $r$ is the keep rate of this layer. 
%Even though we perform the arg-max operation to select the top-k visual tokens according to the attention map during patches selection, we can still backward the gradients for the model during training. As a result, t
The selected image tokens are kept, and the un-selected image tokens are further fused by an inattentive token fusion operation ITF.

The remaining inattentive patch tokens $\{v_{z_1}, v_{z_2}, \cdots,v_{z_{n-k}}\}$ are treated as text-irrelevant tokens. However, the fixed keep rate may remove some useful tokens, so we fuse inattentive tokens to one token $v_{f}$ by a weighted sum operation to supplement attentive ones as follow:
\begin{equation}
    v_{f} = \sum\limits^{n-k}\limits_{i = 1} a_{cls,z_i} \cdot \hat{v}_{z_i}
\end{equation}
After fusing the inattentive patch tokens,  we reconstruct the $j_{th}$ visual sequence as $v^{j} = \left[v^{j}_{cls},v^{j}_{1}, \cdots,v^{j}_{k},{v}^{j}_{f}\right]$, which consists of the image [CLS] token embedding, the selected text-aware image patch embedding, and fused inattentive patch embedding. Then the new visual sequence is fed to the feed-forward network (FFN). 
\begin{figure}
    \centering
    \includegraphics[width=0.9\linewidth]{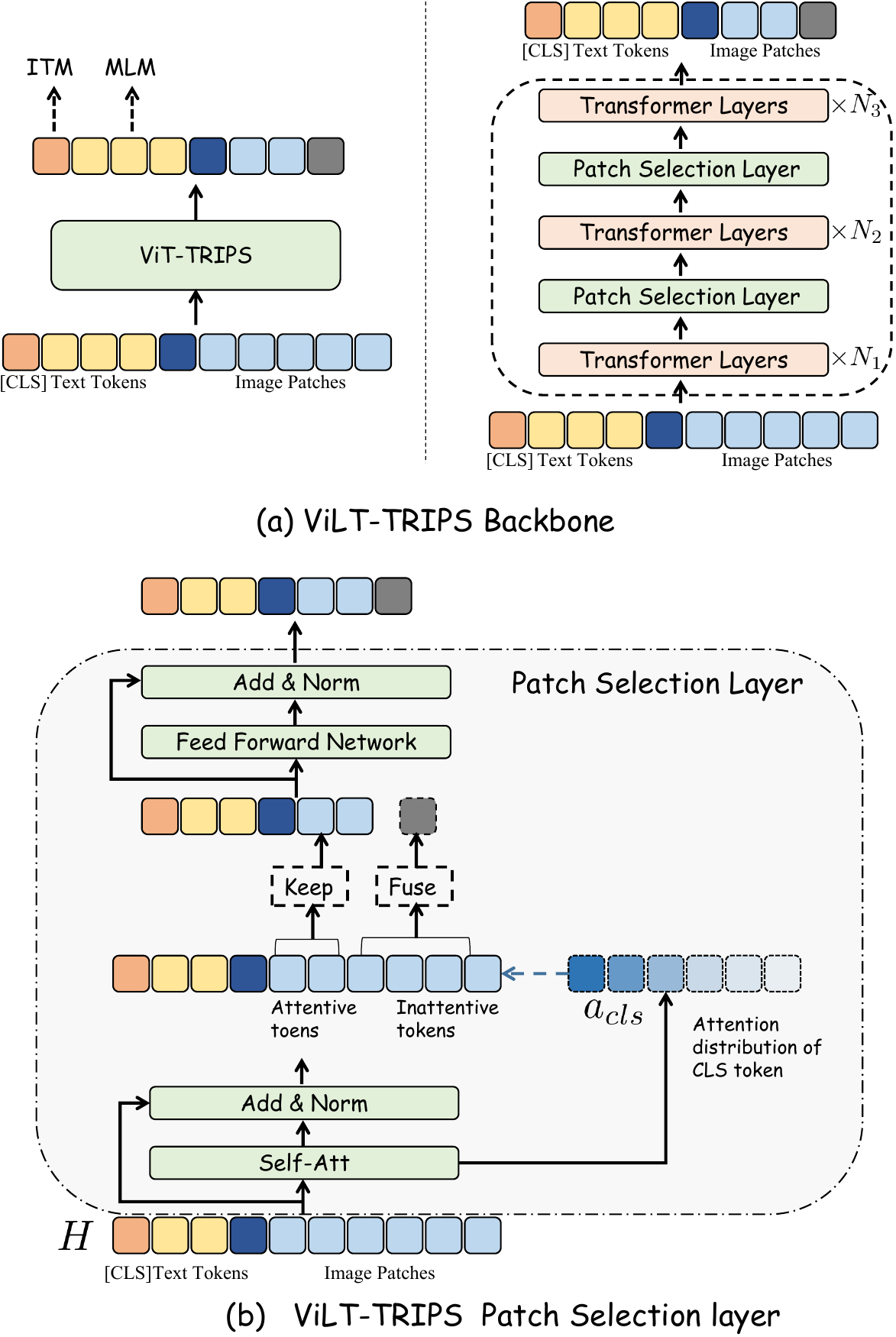}
    \caption{Sub-figure (a) depicts the ViT-TRIPS backbone of the single-stream VLP model. Sub-figure (b) is the illustration of the Text-Relevant Image Patch-Selection layer.}
    \label{fig:vilt-trips}
\end{figure}

\subsection{Extension to Generative VLP Models}

Our method can be easily applied to generative vision-language models. We extend TRIPS to the latest state-of-the-art dual-stream generative model mPLUG~\citep{li2022mplug}, referred to as TRIPS-mPLUG. Specifically, we replace the ViT-based visual backbone in mPLUG with ViT-TRIPS while keeping other settings unchanged. 
% For more details about the model architecture, please refer to the original paper~\citep{li2022mplug}.

\subsection{Extension to Single-stream VLP Models} 

The proposed Text-Relevant Image Patch Selection layer can also be extended to single-stream models, which employ the [CLS] token of the multimodal encoder to preserve attentive image tokens and fuse inattentive tokens to accelerate training and inference. In this paper, we implement this model, denoted as TRIPS-ViLT, based on the ViLT~\citep{Kim2021ViLTVT} framework, a widely used single-stream model. As shown in Figure~\ref{fig:vilt-trips}, the single-stream model has a global [CLS] token for aggregating visual and linguistic information, allowing us to directly select text-aware visual tokens and fuse unrelated image patch tokens based on the attention weights of the [CLS] tokens to image patch tokens.

\subsection{Pre-training Objectives} 
\begin{table}[htbp]
\caption{Pre-training tasks of different TRIPS variants.}
\begin{tabular}{ccccc}
\toprule[1.5pt]
 Model& ITC & ITM & MLM  & PrefixLM \\ \hline
TRIPS-ViLT & \XSolidBrush   & \checkmark & \checkmark & \XSolidBrush \\
 TRIPS-ALBEF& \checkmark & \checkmark & \checkmark & \XSolidBrush  \\
 TRIPS-mPLUG&\checkmark  &\checkmark  & \checkmark &  \checkmark\\ \bottomrule[1.5pt]
\end{tabular}

\label{pre-training data}
\end{table}
We pre-train our models with the following standard objectives: Image-Text Contrastive learning (ITC), Image-Text Matching (ITM), and Masked Language Modeling (MLM). These pre-training tasks are optimized jointly. As shown in Table~\ref{pre-training data}, the three TRIPS variants differ in the combination of pre-training tasks.

\noindent \textbf{Image-Text Contrastive (ITC)} For TRIPS-ALBEF and TRIPS-mPLUG, we follow \citep{Li2021AlignBF} and apply ITC to align the image representation and text representation from the unimodal encoders.

\noindent \textbf{Image-Text Matching (ITM)} The goal of image-text matching is to predict whether the input image and text are matched, which is a shared task among all model variants. We follow the design of \citep{Li2021AlignBF} and select hard negative image-text pairs based on the contrastive text-image similarity. We take the text [CLS] embedding of the multimodal encoder's output as the joint representation, followed by a Multi-Layer Perceptron (MLP) layer for prediction.

\noindent \textbf{Masked Language Modeling (MLM)} The task setup is essentially the same as in BERT~\citep{Devlin2019BERTPO}, where we randomly mask 15\% of tokens in text and the model is asked to predict these masked words with the cross-modal representations.

\noindent \textbf{Prefix Language Modeling (PrefixLM).} This task is specific to mPLUG~\citep{li2022mplug}, which aims to generate a caption given an image and predict the text segment subsequent to the cross-modal context as in ~\citep{bi2020palm}. It optimizes a cross-entropy loss by maximizing the likelihood of text in an autoregressive manner.

\section{Experiment Settings}
\subsection{Implementation Details}
We pre-train TRIPS-ALBEF and TRIPS-mPLUG for 30 epochs, and TRIPS-ViLT for 20 epochs with a total batch size of 512 on 8 NVIDIA V100 GPUs. We initialize the visual encoder using CLIP (ViT-B/16) \citep{Radford2021LearningTV} pre-trained on 400M noisy image-text pairs and employ the AdamW \citep{Loshchilov2019DecoupledWD} optimizer with a weight decay of 1e-2. The learning rate is warmed up to 1e-5 (ViT-B/16) and 1e-4 (BERT${base}$) in the first 1000 iterations. During pre-training, we use input images with a resolution of $256 \times 256$ and increase the image resolution during fine-tuning. For TRIPS-ALBEF and TRIPS-mPLUG, we utilize a 6-layer Transformer for both the text encoder and the cross-modal fusion network. Following \cite{Li2021AlignBF}, we initialize the text encoder with the first 6 layers of BERT${base}$~\citep{Devlin2019BERTPO} and the cross-modal network with the last 6 layers of BERT$_{base}$. For image-text contrastive learning, we set the queue size to 65,536 and the momentum coefficient to 0.995.

In the proposed TRIPS-ViT backbone, we designate the 5th and 10th layers as patch-selection layers and set the keep rate for each layer to 70\%, achieving a balance between downstream task performance and model inference speed. Details of fine-tuning hyperparameters can be found in Appendix \ref{sup:downstream task details}.

We use the AdamW optimizer~\citep{Loshchilov2019DecoupledWD} with a weight decay of 0.02 to train our models. For TRIPS-ALBEF and TRIPS-mPLUG, we warm up the learning rate to 1e-5 (ViT-B/16) and 1e-4 (BERT$_{base}$) in the first 1000 iterations, then decay it to 1e-6 according to a cosine schedule. Pre-training TRIPS-ALBEF takes approximately 60 hours, TRIPS-mPLUG takes around 80 hours, and TRIPS-ViLT requires about 40 hours; all performed on 8 V100-32G GPUs using a 4M pre-training dataset for approximately 20 epochs.

To enhance the generalization of vision encoders during pre-training, we apply RandAugment~\citep{cubuk2020randaugment} to random image crops of size 256 $\times$ 256. In the fine-tuning stage for VQA, image captioning, and visual grounding tasks, we increase the image resolution. For image-text contrastive learning, we set the queue size to 65,536 and the momentum coefficient to 0.995.

\subsection{pre-training data}

\label{sup:pre-training data}
\begin{table}[htbp]
\setlength\tabcolsep{10pt}
\centering
\caption{Statistics of the pre-training datasets.}
\begin{tabular}{l|cccc}

\toprule[1.5pt]
  &  COCO & VG & SBU & CC3M \\
\midrule
image & 113K & 100K & 860K & 3M   \\
text & 567K & 769K & 860K & 3M \\
\bottomrule[1.5pt]
\end{tabular} 

\label{table:pretraindata}
\end{table}

% \begin{table}[htbp]
% \setlength\tabcolsep{7pt}
% \centering

% \begin{tabular}{l|cccc}
% \toprule[1.5pt]
%      &  image & Captions & Objects & Regions \\
% \midrule
% COCO & 0.11M & 0.55M & 0.45M & -   \\
% VG   & 0.10M & - & 2.0M & 3.7M \\
% \bottomrule[1.5pt]
% \end{tabular} 
% \caption{Statistics of objects/regions annotations used in the pre-training.}
% \label{table:objectdata}

% \end{table}
% Table \ref{table:pretraindata} shows the statistics of the 4M images with texts used in the pre-training stage. Besides, As shown in Table~\ref{table:objectdata} we use also use the objects/regions annotions from COCO\citep{Lin2014MicrosoftCC} and VG \citep{Krishna2016VisualGC} datasets and we give a statistics of object and region annotations of each dataset. Note that we use the object/region annotations provide by ~\citep{Zeng2021xvlm} thus we following their setting which filtered out some samples because of: 1) invalid annotations (e.g. negative values for bounding boxes or boxes being outside of the images); 2) boxes being too small (< 1\%); 3) highly overlapped textual descriptions of regions (>75\%), etc. After pre-processing, we keep: for example, COCO objects 446,873 (from 859,999), VG objects 2,043,927 (from 3,802,349), VG regions 3,699,598 (from 5,402,953).
We construct our pre-training data using two web datasets (Conceptual Captions \citep{Sharma2018ConceptualCA}, SBU Captions \citep{Ordonez2011Im2TextDI}) and two in-domain datasets (MSCOCO \citep{Lin2014MicrosoftCC} and Visual Genome \citep{Krishna2016VisualGC}). The total number of unique images is 4.0M, and the number of image-text pairs is 5.1M.
 Table \ref{table:pretraindata} shows the statistics of the 4M images with texts used in the pre-training stage.

\section{Experiment Results}

\subsection{Overall Performance}
We evaluate TRIPS-ALBEF, TRIPS-mPLUG, and TRIPS-ViLT on five vision-language downstream tasks: visual question answering (VQA), natural language for visual reasoning (NLVR), image-text retrieval, image captioning, and visual grounding. 
Our baselines cover 15 VLP models, which is detailed in Section~\ref{sup:comparison models}. We will first analyze their overall performances on these tasks. 

\subsubsection{Visual Question Answering}

The VQA task \citep{Agrawal2015VQAVQ} requires the model to answer natural language questions given an image. Following \citep{Li2021AlignBF}, we treat VQA as an answer generation problem. We report test-dev and test-std scores by submitting our results to the evaluation server\footnote{https://eval.ai/web/challenges/challenge-page/830/overview}, as shown in Table \ref{table:vqa_caption}. Looking into Table~\ref{table:vqa_caption}, VLP models equipped with ViT-TRIPS improve performance on the VQA task (e.g., ViLT~\citep{Kim2021ViLTVT} achieves 71.26 on Test-dev, while TRIPS-ViLT scores 71.48; ALBEF \citep{Li2021AlignBF} reaches 76.12, while TRIPS-ALBEF attains 76.23). These results are impressive, considering they are achieved with significantly reduced training and inference costs (see more details of efficiency in Section~\ref{efficiency}).

\begin{table}[htpb]
\centering
\caption{Evaluation Results on VQA and NLVR$^2$. More details about comparison models are in Appendix \ref{sup:comparison models}. } 
\setlength \tabcolsep{1.0pt}
\begin{tabular}{l|c|cc|cc}

\toprule[1.5pt]
\multirow{2}{*}{model} & \multirow{2}{*}{\begin{tabular}[c]{@{}c@{}}\# Pre-train\\ Data \end{tabular}} & \multicolumn{2}{c|}{VQA} & \multicolumn{2}{c}{NLVR2} \\
                       &                                                                                   & Test-dev   & Test-std   & Dev        & Test-P       \\ \hline
 E2E-VLP& 4M & 73.25  & 73.67 &    77.25 & 77.96      \\
OSCAR$_{Base}$ & 6.5M & 73.16 & 73.44 &  78.07 & 78.36 \\
VinVL$_{Large}$ & 5.65M & 76.52 & 76.60 & 82.67 & 83.98 \\
ViLBERT  & 3.3M  & 70.63 & 70.92 & - & - \\
VisualBERT  & 180K & 70.80 & 71.00&67.40 & 67.00\\
LXMERT & 180K & 72.42 & 72.54 & 74.90 & 74.50\\
UNITER$_{Large}$& 4M & 73.82 & 74.02 & 79.12 & 79.98 \\
% SimVLMhuge_{huge} \citep{wang2021simvlm} & 1.8B & 40.6 & 33.7 & 143.3 & 25.4 & - & - & - & - & 112.2 & -  \\
% LEMONlarge_{large} & 200M & - & - & 40.6 & 30.4 & 135.7 & 23.5 & 42.3 & 31.2 & 144.3 & 25.3 & 113.4 & \textbf{15.0} \\
METER$_{CLIP}$ & 4M & 77.68 & 77.64  & 82.33 & 83.05 \\
BLIP$_{Base}$ & 14M & 77.54 & 77.62 &  82.67 & 82.30\\
VLMo$_{Large}$ & 4M &76.64 & 76.89  & 82.77 & 83.34 \\
SimVLM$_{Base}$ & 1.8B & 77.87 & 78.14  & 81.72 & 81.77  \\ \hdashline
ViLT$_{Base}$ & 4M  & 71.26 & 71.29  & 75.18  & 76.2 \\
TRIPS-ViLT  & 4M &  71.48 & 71.52  & 75.89 & 76.4 \\   \hdashline
% OFA  & 18M & 79.87 & 80.02 & - & - & - & - & 43.5 & 31.9 & 149.6 & \textbf{26.1} & - & - \\
% XVLM~\citep{Zeng2021xvlm} & 4M & 78.07 & 78.09 & 39.8 & - & 133.1 & -& 41.3 & - & 140.8 & -  & 84.16 & 84.21 \\ \midrule
ALBEF$_{Base}$ & 4M  &  76.12     & 76.32 & 82.21 & 83.1 \\
TRIPS-ALBEF & 4M &  76.23 & 76.48  & 82.35      &     83.34   \\ \hdashline
mPLUGG$_{Base}$ & 4M & 77.80  & 77.98  & 83.76 & 83.92 \\
TRIPS-mPLUG & 4M & \textbf{77.89}  & \textbf{78.23}& \textbf{83.87} & \textbf{84.04}\\
\bottomrule[1.5pt]
\end{tabular}

\label{table:vqa_caption}
\end{table}
\subsubsection{Natural Language for Visual Reasoning}

The NLVR2 \citep{Suhr2019ACF} task requires the model to predict whether a sentence accurately describes a pair of images, which is a binary classification task. For TRIPS-ALBEF and TRIPS-mPLUG, we follow \citep{Li2021AlignBF} and use two cross-attention layers to process the two input images; their outputs are merged and fed into a Feed Forward Network (FFN). An MLP classifier is then applied to the output embedding of the text [CLS] token. For TRIPS-ViLT, we follow \citep{Kim2021ViLTVT} and use the pairing method. The input is reformulated into two pairs (question, image1) and (question, image2), and each pair goes through TRIPS-ViLT. The classification head takes the concatenation of two pooled representations as input and outputs the binary prediction. As shown in Table~\ref{table:vqa_caption}, TRIPS outperforms existing VLP methods.

\begin{table*}[htbp]
\caption{Evaluation results of image-text retrieval on Flickr30K~\citep{Plummer2015Flickr30kEC} and COCO datasets~\citep{Lin2014MicrosoftCC}.}
\setlength\tabcolsep{3.3pt}
\centering
\small
\begin{tabular}{l|c|cccccc|cccccc}
\toprule[1.5pt]
\multicolumn{1}{c|}{\multirow{2}{*}{Models}}      &
%\multirow{2}{*}{\# Pre-train} &
\multicolumn{1}{c|}{\# Pre-train} &
\multicolumn{6}{c|}{MSCOCO (5K test set)} & \multicolumn{6}{c}{Flickr30K (1K test set)} \\
      &  data & \multicolumn{3}{c}{TR} & \multicolumn{3}{c|}{IR} & \multicolumn{3}{c}{TR} & \multicolumn{3}{c}{IR}          \\
\midrule
&&R@1&R@5&R@10&R@1&R@5&R@10&R@1&R@5&R@10&R@1&R@5&R@10 \\ \hline
E2E-VLP & 4M     &-& -&-&-&-&- & 86.2 &97.5 &98.92&73.6 & 92.4 &96.0 \\
OSCAR$_{Base}$ & 4M  & 70.0&91.1&95.5&54.0&80.8&88.5&-& -&-&-&-&-   \\
VinVL$_{Large}$ & 5.65M & 75.4 & 92.9 & 96.2 &58.8 &83.5 &90.3 &-& -&-&-&-&-   \\
ViLBert & 3.3M  & - &- &- &- &- &-s& -& -&-&58.2 & 84.9 & 91.5  \\
UNITER$_{Large}$ & 4M     & 65.7&88.6&93.8&52.9&79.9&88.0&87.3& 98.0&99.2&75.6&94.1&96.8  \\
METER$_{CLIP}$ & 4M & 76.2 & 93.2 & 96.8& 57.1 & 82.7 & 90.1 &94.3 & 99.6 & 99.9  &82.2  &96.3 & 98.4  \\

VLMo$_{Lagre}$ & 4M & 78.2& 94.4& 97.4& 60.6& 84.4& 91.0& 95.3& 99.9& 100.0& 84.5& 97.3& 98.6 \\
% Florence & 0.9B & 81.8&95.2&-&63.2&85.7&-&97.2& 99.9&-&87.9&98.1&-                 \\
BLIP$_{Base}$ & 14M & 80.6 &95.2&97.6&63.1&85.3&91.1&96.6& 99.8&100.0&\textbf{87.2}&97.5&98.8                 \\
\hdashline
% TRIPS & 4M  & 80.6 &95.2&97.6&63.1&85.3&91.1&96.6& 99.8&100.0&87.2&97.5&98.8  
ViLT & 4M  & 61.5 & 86.3 &92.7& 42.7& 72.9&83.1&83.5& 96.7 &98.6 &64.4 &88.7&93.8\\
TRIPS-ViLT & 4M  & 63.2 & 88.2 &94.1& 43.9& 74.1&84.6&85.4 & 98.3 &98.4& 64.4 &89.2&94.9  \\\hdashline
ALBEF$_{Base}$ & 4M & 77.8&94.3&97.4&60.3& 84.7&91.0&95.9& 99.8&100.0&85.6&97.5& 98.9                 \\
TRIPS-ALBEF & 4M  & 78.1 &94.8 &97.6&61.3&84.3&91.4&96.3&99.8&100.0& 85.8 &98.1 & 99.0   \\ \hdashline
mPLUG$_{Base}$& 4M  & 80.5 &95.4&97.9&63.3&85.3&91.2&96.7& 99.8&100.0& 86.5&97.5& 98.9   \\
TRIPS-mPLUG & 4M  & \textbf{80.8} & \textbf{95.7} & \textbf{98.0} & \textbf{63.6} & \textbf{85.5} & \textbf{91.5}& \textbf{97.0}& 99.8&100.0&86.9 & \textbf{97.8} & \textbf{99.1}   \\
\bottomrule[1.5pt]
\end{tabular}      

\label{table:retrieval}

\end{table*}

\begin{table*}[htbp]
\centering
\caption{Evaluation Results on image captioning on COCO Karpathy test split \citep{karpathy2015deep}. B@4: BLEU@4, M: METEOR, C: CIDEr, S: SPICE. }
\setlength\tabcolsep{7.5pt}
\begin{tabular}{l|c|cccccccc}
\toprule[1.5pt]
\multirow{3}{*}{Models} & \multirow{3}{*}{\begin{tabular}[c]{@{}c@{}}\# Pre-train \\ Data\end{tabular}} & \multicolumn{8}{c}{COCO Caption}                                                        \\
                        &                                                                              & \multicolumn{4}{c}{Cross-entropy Optimization} & \multicolumn{4}{c}{CIDEr Optimization} \\
                        &                                                                              & B@4         & M         & C         & S        & B@4       & M       & C       & S \\  \midrule     
E2E-VLP & 4M  & 36.2 &-&117.3&-&  - & - & - & - \\
OSCAR$_{Base}$& 6.5M & 36.5 & 30.3 & 123.7 & 23.1 & 40.5 & 29.7 & 137.6 & 22.8  \\
VinVL$_{Large}$ & 5.65M  & 38.5 & 30.4 & 130.8 & 23.4 & 41.0 & \textbf{31.1} & 140.9 & 25.2  \\
BLIP$_{Base}$& 14M & 38.6 & - & 129.7 & - & - & - & - & - \\
SimVLM$_{Base}$& 1.8B & 39.0 & \textbf{32.9} & \textbf{134.8} & \textbf{24.0} & - & - & - & -  \\ \hdashline
mPLUGG$_{Base}$ & 4M  & 39.3 & 30.1 & 132.4 & 23.34 & 41.2 & 31.0 & 140.8 & 25.2  \\
TRIPS-mPLUG & 4M & \textbf{39.4} & 30.5 & 132.8 & 23.82 & \textbf{41.5} & 30.9 & \textbf{141.2} & \textbf{25.4} \\
\bottomrule[1.5pt]
\end{tabular}

\label{table:captioning}
\end{table*}
\subsubsection{Image-Text Retrieval}

We conduct experiments for both image-to-text retrieval (TR) and text-to-image retrieval (IR) on MSCOCO \citep{Lin2014MicrosoftCC} and Flickr30K \citep{Plummer2015Flickr30kEC} datasets. We jointly optimize the ITC loss and the ITM loss during fine-tuning. Results are reported in Table \ref{table:retrieval}. As illustrated in Table \ref{table:retrieval}, our model achieves comparable performance to other VLP baselines.

\subsubsection{Image Captioning} 
\vspace{-1ex}
% As there is no textual input in the image caption task, we directly set the hyperparameter $\beta$ to 0 and select the patches based on the attention weight of the image [CLS] token to other image tokens. Following~\citep{Li2020OscarOA}, we first fine-tune TRIPS with cross-entropy loss and then with CIDEr optimization~\citep{scst} for extra 5 epochs. As shown in Table~\ref{table:vqa_caption}, when we set the image resolution to $384 \times 384$, TRIPS can still get the comparable result with SOTA models, including XVLM and BILIP on both COCO Caption and Nocaps datasets. When we set  the image resolution to $512 \times 512$.TRIPS performs the best on CIDEr evaluation and surpasses the SOTA model.

To investigate the text generation capacity of TRIPS-mPLUG, we evaluate it on the image captioning task. Since there is no textual input for this task, we select visual semantic-aware patches based on the attention weight of the image [CLS] token to other image tokens and fuse other image tokens based on the attention weight. We perform this operation based on the observation that the image [CLS] token in ViTs pays more attention (i.e., having a larger attention value) to class-specific tokens than to tokens on non-object regions~\citep{Caron2021EmergingPI}. Following~\cite{Li2020OscarOA}, we fine-tune TRIPS-mPLUG with cross-entropy loss and then with CIDEr optimization for an additional 5 epochs. Our experiments, as shown in Table~\ref{table:captioning}, demonstrate that TRIPS-mPLUG achieves comparable results with state-of-the-art models.

\subsubsection{Visual Grounding} 

\begin{table}[htbp]
\centering
\caption{Evaluation results of visual grounding on ReferCOCO+. We use the accuracy of IOU 0.5 on visual grounding (a prediction is right if the IoU between the grounding-truth box and the predicted bounding box is larger than 0.5)}
 \setlength{\tabcolsep}{4mm}{
\begin{tabular}{@{}lccccc@{}}
\toprule[1.5pt]
\multicolumn{1}{c}{\multirow{2}{*}{Model}} & \multicolumn{3}{c}{RefCOCO+} \\
\multicolumn{1}{c}{}          & testA   & testB   & val          \\ \midrule

UNITER$_{Large}$  & 75.90    & 81.45   & 66.70         \\
% VL-BERT~\citep{Su2020VLBERTPO} &72.59  & 78.57 & 62.30   \\
ViLBERT  & 72.34&  78.52 &   62.61             \\
VILLA & 76.17    & 81.54   & 66.84            \\
MDETR   & 79.52    & 84.09   & 70.62      \\
UNICORN &\textbf{ 80.30}    & 85.05   & \textbf{71.88}      \\\hdashline
% XVLM~\citep{Zeng2021xvlm}     & 80.17  &  86.36  & 71.00 \\ 
% ALBEF$_{Base}$~\citep{Li2021AlignBF}    & 78.23  &  80.12   & 67.84         \\
% TRIPS-ALBEF &78.34  &  80.78  & 68.42     \\ \hdashline
mPLUG$_{Base}$   & 80.07  &   85.21 &  71.03       \\
TRIPS-mPLUG & 80.11  &   \textbf{86.03} &  71.21      \\ 
\bottomrule[1.5pt]
\end{tabular}}

\label{tab:visual_grounding}
 
\end{table}

% Table \ref{tab:visual_grounding} shows that when we set the image resolution to $384 \times 384$, TRIPS get the comparable result compared with SOTA methods.  Due to the elimination of the undetected redundant patches in the visual backbone of the model, we can improve the image resolution and get a better result without increasing more computational costs compared with other methods. When we increase the image resolution to $512 \times 512$, our model outperforms all the SOTA methods, which indicates the efficiency and effectiveness of TRIPS.

Following the setting of mPLUG~\citep{li2022mplug}, we also evaluate TRIPS-mPLUG on the visual grounding task, which requires models to localize the referred object in the image based on a given text query. Instead of directly regressing the bounding boxes, we concatenate visual features and attended textual features and feed them into the decoder to predict the coordinates. Table \ref{tab:visual_grounding} demonstrates the performance of TRIPS-mPLUG in the visual grounding task. TRIPS-mPLUG achieves comparable results with competitive baseline methods.

\begin{figure}[t!]
     \centering
     \includegraphics[width=0.48\textwidth]{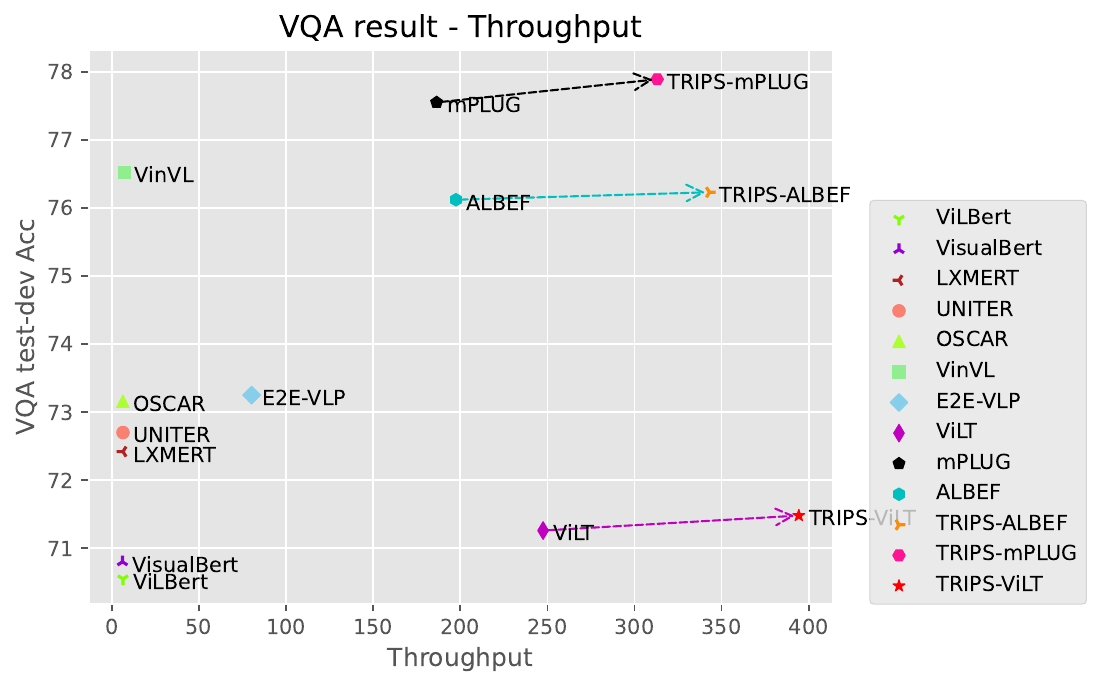}
     \caption{The figure below visualizes the performance and inference speed distribution of different VLP models. The y-axis represents the accuracy of the VLP model on the test-dev dataset of VQA, while the x-axis represents the throughput of processing image-text pairs on a single 32G V100 GPU.}
     \label{fig:efficiency}
     \vspace{-3ex}
\end{figure}

\begin{table}[!t]
\small
\caption{The comparison of the efficiency of different models. FLOPs, throughput, and latency are reported here. Since FLOPs are proportional to input size, for a fair comparison, we use same the input size with \citep{Kim2021ViLTVT}, which is 197 for image patches length and 40 for text tokens length. We also keep the same setting when calculating throughput and latency.}
\setlength{\tabcolsep}{1.7mm}
\begin{tabular}{lccc}
\toprule[1.5pt]
Models & Latency &  FLOPs  & Throughput   \\ \hline
ViLBERT & 880ms &   958.10 &  6.40  \\
VisualBERT & 910ms  &425.20 &  6.10 \\
LXMERT   &  980ms  &952.20 &  6.38  \\
UNITER$_{Base}$   & 870ms &     949.90    &   6.42      \\

OSCAR$_{Base}$ & 860ms &     956.40  &    6.35          \\
VinVL$_{Base}$  & 640ms &    1023.30    &    7.32             \\
E2E-VLP   &  70ms&    144.30     &     80.23        \\ \hdashline
ViLT$_{Base}$     &  15ms&     55.40   &  247.53          \\ 
TRIPS-ViLT    & \textbf{8}ms&  32.38  &      \textbf{394.32}      \\  \hdashline
ALBEF$_{Base}$     & 22ms&  33.42    &      197.52     \\
TRIPS-ALBEF & 11ms & \textbf{ 20.89 } &  343.05    \\   \hdashline
mPLUG$_{Base}$   & 24ms&  36.63    &      186.42      \\
TRIPS-mPLUG   & 13ms&   25.04  &    313.71     \\   

\bottomrule[1.5pt]
\end{tabular}
\centering

\label{table_efficient}

\end{table}

\subsection{Efficiency of TRIPS} \label{efficiency}

To investigate the efficiency of Text-Relevant Image Patch Selection, we first compare the computational complexity of various models. We report the Floating Point Operations Per Second (FLOPs), a widely used evaluation metric for model computational complexity. In addition, we evaluate the computational speed of our model by comparing the throughput and latency of different models. We use a Xeon Platinum 8163 CPU and an NVIDIA V100 GPU to calculate the latency and throughput. As shown in Table \ref{table_efficient}, TRIPS not only exhibits the lowest computational complexity but also achieves the fastest computational speed (e.g., 343.05 throughput and 11ms latency for TRIPS-ALBEF vs. 197.52 throughput and 22ms latency for ALBEF).

To provide a more intuitive illustration of how TRIPS balances model efficiency and effectiveness, we visualize the performance and inference speed distribution of different VLP models on the VQA task in Figure \ref{fig:efficiency}. The figure clearly demonstrates that incorporating TRIPS consistently results in a speedup while maintaining competitive or superior performance across various VLP base models.

Similar to this visualization, the following further analyses will mainly be based on the VQA task to make the tables and figures more reader-friendly. Note that most conclusions apply to other tasks.

\begin{table*}[h]
\caption{Results of pre-training and finetuning TRIPS-ALBEF with different locations and keep rates. we report the text-dev score results of VQA, FLOPs and Throughput. In this table, we set the input image size to 384 $\times$ 384 and the length of input text is 40. The throughput (image-text/s) is measured on an NVIDIA V100 GPU using the largest possible batch size for our model. }
 \setlength{\tabcolsep}{3.5mm}{
\begin{tabular}{ccccccc}
\toprule[1.5pt]
 Locations & Keep rates    & Overall Keep rate & VQA test-dev &  FLOPs (G) & Throughput \\ \hline
 -                                                                  & -            & 100\%    &   76.12 &   76.03       & 79.32  \\ \hline
{[}2{]}                                                            & 50\%             & 50\% &   75.60  &  42.17    &  161.26 \\
{[}10{]}                                                           &50\%              &  50\% &  76.19  &    63.84   & 96.66  \\ \hline
 {[}2,4{]}                                                          & 50\%     &   \%25 & 74.21  &  \textbf{28.00}    &  \textbf{238.41} \\
{[}4,8{]}                                                          & 50\%    &   \%25 &  74.93 &   38.82  &  165.30 \\
{[}5,10{]}                                                          & 50\%    &   \%25 & 75.29 &  44.22  & 143.37 \\
{[}6,12{]}                                                        & 50\%     &   \%25 &  75.48 &   49.63  & 126.46  \\\hline
{[}2,4{]}                                                          & 70\%    &  \%49 &  74.87 &   43.96   &  153.92\\
{[}4,8{]}                                                          & 70\%     &   \%49 &  75.94 &   51.72  &  125.55 \\
{[}5,10{]}                                                          & 70\%     &   \%49 & 76.23&  55.60  &  115.01 \\
{[}6,12{]}                                                        & 70\%     &   \%49 &   \textbf{76.24}&   59.48  &  106.07 \\ \hline
{[}2,6,10{]}                                                        & 70\%  &   \%34 & 74.92 &   42.66  & 156.13\\
{[}3,6,9{]}                                                        & 70\%  &   \%34 &  75.09  & 43.49 &   151.40 \\
{[}4,8,12{]}                                                       & 70\%  &   \%34 &   75.23 & 49.74  & 129.81 \\ \bottomrule[1.5pt]
\end{tabular}}
\centering

\label{table4}
\end{table*}
\begin{table*}[h]
\caption{Results of TRIPS-ALBEF finetuning on VQA and NLVR task with different resolution images. When calculating FLOPs, the input length of the text is kept at 40 and the settings for calculating throughput are the same as Table\ref{table4}.}
\setlength{\tabcolsep}{5.0mm}{

\begin{tabular}{@{}cccccc@{}}
\toprule[1.5pt]
 Selection Layer         & Keep rate     & image size     & VQA test-dev  & FLOPs(G) & Troughout \\ \midrule
 - & - & $384 \times 384$ &  76.12    & 76.03   & 79.32 \\ \hline
 {[}5,10 {]}            & 70\% & $224\times224$ &     75.23    &\textbf{ 20.89} &  \textbf{ 343.05}   \\
 {[}5,10 {]}            & 70\% & $256\times256$ &      75.84   &    26.61  &  258.03     \\
 {[}5,10 {]}            & 70\% & $304\times304$ &    76.13    &    36.62   &  189.07\\
 {[}5,10 {]}            & 70\% & $384\times384$ &      76.23    &    55.60 &  115.01    \\
  {[}5,10 {]}            & 70\% & $464\times464$ &    76.54   &  74.83  &  81.0   \\
 {[}5,10 {]}            & 70\%& $512\times512$ &       \textbf{  76.83}   & 84.13   & 72.10      \\ \bottomrule[1.5pt]
\end{tabular}}
\centering

\label{table5}
\end{table*}

\subsection{ The Impact of Patch-section Location and Keep Rate }

To assess the impact of patch-selection layer location and keep rate on the model's efficiency and effectiveness, we train TRIPS-ALBEF with varying patch-selection locations and numbers of selected tokens, and test the yielding models on the VQA task. The results presented in Table \ref{table4} reveal two key observations. 

First, positioning the patch-selection layer in shallower layers reduces computational complexity but negatively affects accuracy. For instance, when the patch-selection layer is placed before the third layer (i.e., at the second layer), accuracy significantly drops despite a notable increase in throughput. A possible explanation for this is that the attention maps between the text [CLS] embedding and the input tokens may be unreliable during the early processing of input tokens in shallow layers. 

Second, fusing too many image tokens in the patch-selection layer can lead to a considerable decline in downstream task performance. For example, if we designate the 2nd and 4th layers in ViT as patch-selection layers and set the keep rate to 50\%, performance on the VQA task decreases to 74.21, compared to 76.12 for the model without a patch-selection layer.

\begin{figure*}[htbp]
\centering
\subfigure[VQA results of Different Resolutions]{
\includegraphics[width=0.33\textwidth]{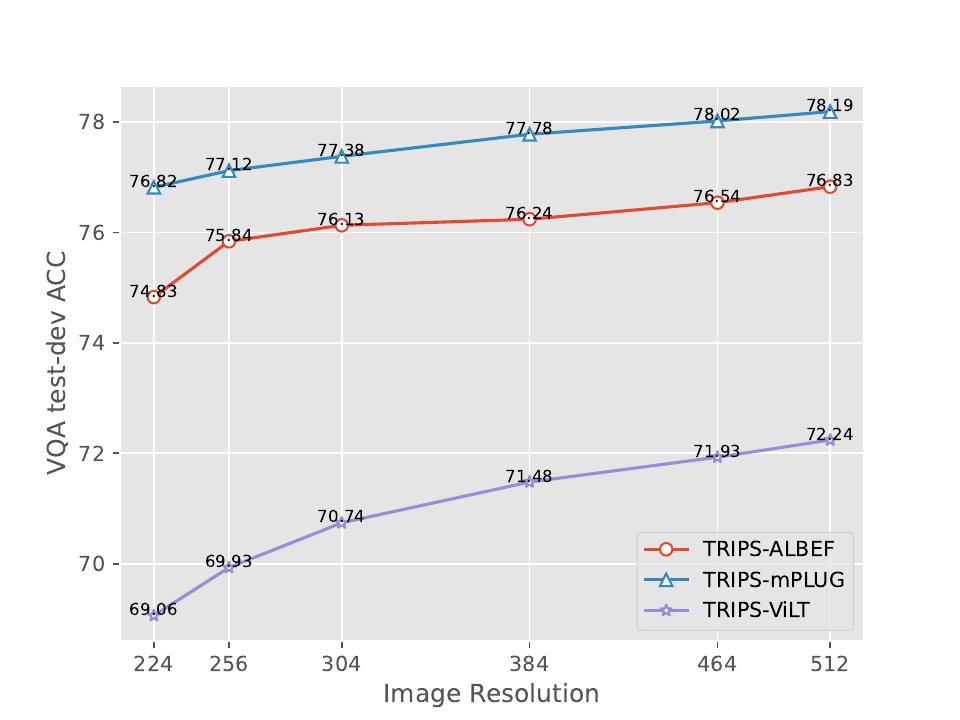}
% \caption{ROUGE-1}
}%
\subfigure[Throughput of Different Resolutions]{
\centering
\includegraphics[width=0.33\textwidth]{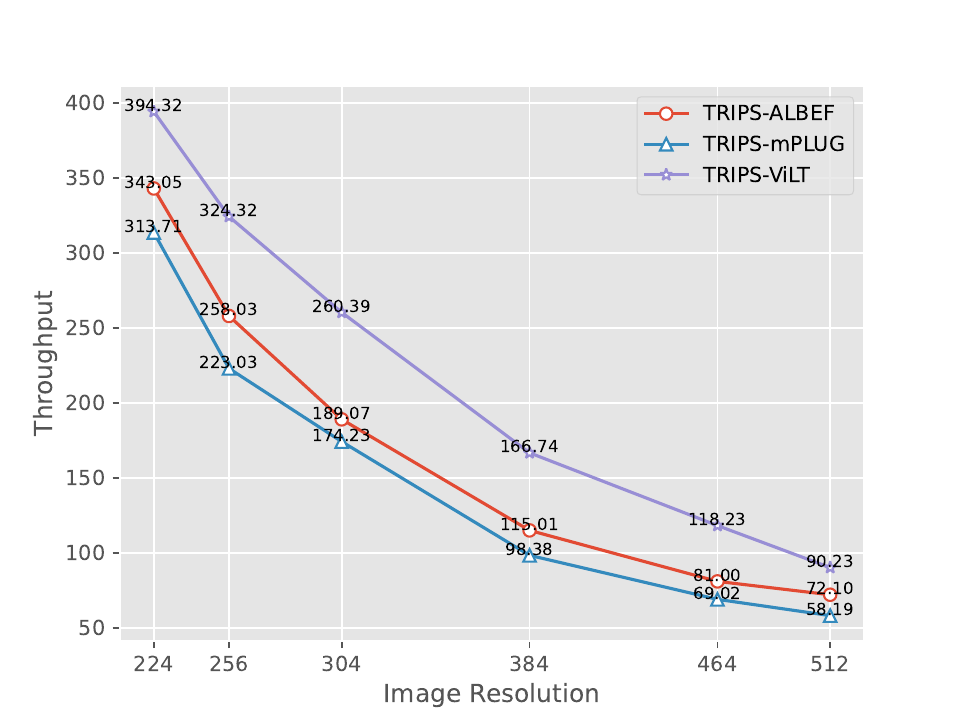}
% \caption{ROUGE-2}
}%
\subfigure[FLOPs of Different Resolutions]{
\centering
\includegraphics[width=0.33\textwidth]{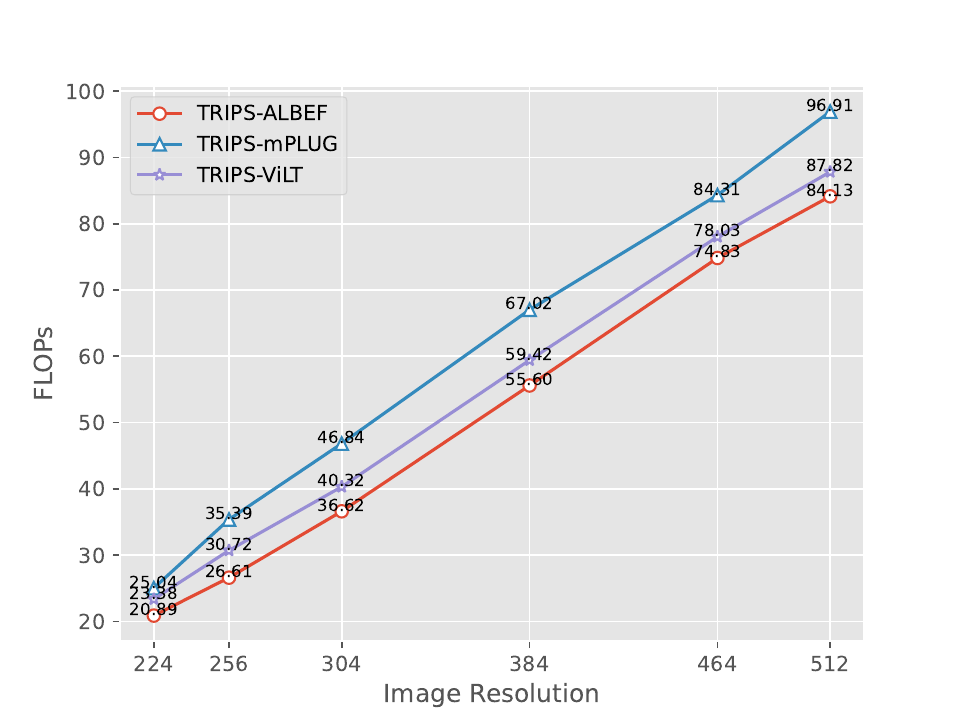}
% \caption{ROUGE-3}
}%

\caption{Sub-figure (a) visualizes the VQA results of three models (TRIPS-ViLT, TRIPS-ALBEF, TRIPS-mPLUG) at different image resolutions, sub-figure (b) visualizes the throughput of three different models at different image resolutions, and sub-figure (c) visualizes the FLOPs of the three models at different image resolutions.}
\label{fig:image_resolution}
 
\end{figure*}

\subsection{Fine-tuning on Higher Resolution Images}

We can manipulate the computational cost by fusing varying numbers of inattentive tokens. To demonstrate this, we fine-tune TRIPS-ALBEF, TRIPS-ViLT, and TRIPS-mPLUG on the VQA task, which uses images with different resolutions as input. The results are presented in Table \ref{table5} and Figure \ref{fig:image_resolution}. These experimental findings indicate that by increasing the input image resolution, we can enhance the model's performance by incorporating more image tokens. For instance, by fine-tuning TRIPS-ALBEF with images of size 464x464, we can achieve a score of 76.54 on VQA, surpassing the baseline fine-tuned with images of 384x384 while maintaining a similar computational complexity.

% \subsection{Effectiveness of Text-Relevant Image Patch Selection }
% To verify the effectiveness of Text-Relevant Image Patch Selection, we first implement the single-stream model TRIPS-S as we present in subsection \ref{sect2}. Then, we examine the downstream task performance, computational complexity, and inference speed of TRIPS and TRIPS-S (both with and without Text-Relevant Image Patch Selection). The results are shown in Figure \ref{fig:image_resolution}, and we find that for both TRIPS and TRIPS-S, we can see a consistent improvement in the inference speed and downstream task performance by incorporating the text-relevant image patch selection mechanism. These results suggest that the proposed image patch selection mechanism is not only efficient but also effective. Notably,  compared with the dual-stream model TRIPS, TRIPS-S is faster in inference due to the parameter efficiency of the single-stream model. However, its performance lags behind state-of-the-art performance on downstream VL tasks.

\begin{table}[h]
\setlength{\tabcolsep}{2.0mm}{
\centering
\caption{The result of ablations. We finetune TRIPS-ViLT, TRIPS-ABLEF, and TRIPS-mPLUG on VQA and report test-dev results. The setting for calculating FLOPs and throughput is the same as Table \ref{table4}. The same with the settings of TRIPS we present in the main results, we select the 5th and 10th as the patch selection layer, and each layer will keep 70\% image tokens. }
\begin{tabular}{@{}ccccc@{}}
\toprule[1.5pt]
model    & VQA & FLOPs(G) & Throughput \\ \midrule
ALBEF-TRIPS     &  76.23   & 55.60   &  115.01        \\  
  -$w/o$ ITF &  75.92 &    55.14  &     118.04     \\ 
  -$w/o$ TD-ATT &   75.23      &  53.44     &     120.23    \\ \hdashline
ALBEF-mPLUG     &  77.89   & 67.02   &  98.38        \\  
  -$w/o$ ITF &   77.29 & 65.23   &  103.38  \\ 
  -$w/o$ TD-ATT &   76.89   &  63.34   &  109.38      \\ \hdashline
ALBEF-ViLT     &  71.48   & 59.42  &  166.74       \\  
  -$w/o$ ITF &  70.63 &  57.19  &  172.01    \\       \bottomrule[1.5pt]
\end{tabular}
}

\label{table6}

\end{table}

\begin{figure}[h]
 
\centering
\includegraphics[width=0.45\textwidth]{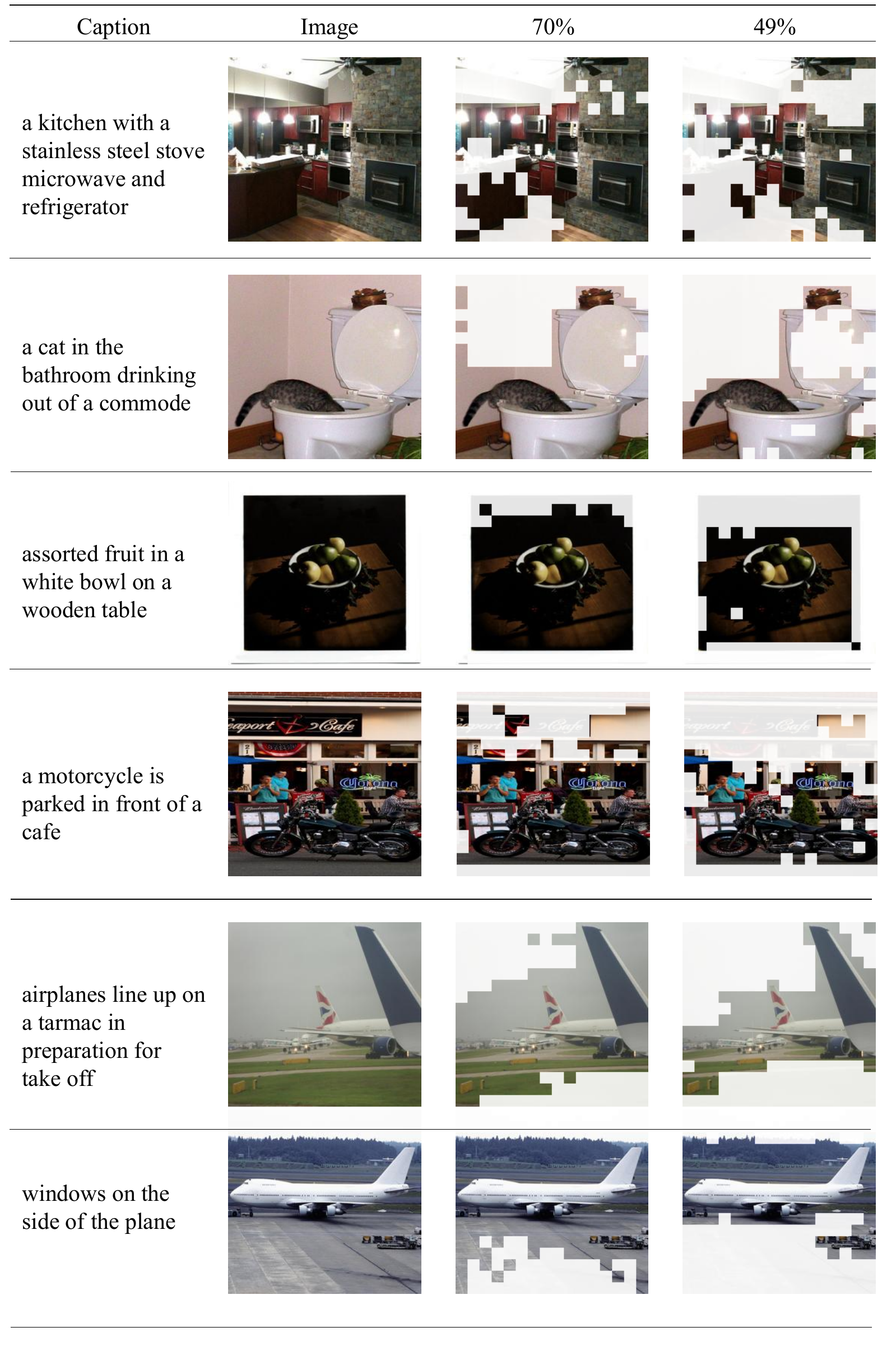}%fig2文件夹下的xbee.esp图片，
\caption{The visualization of the selected text-aware image patches in different selection layers of TRIPS-mPLUG. We set the 5th and 10th layers in the vision backbone as the patch-selection layer, and we keep 70\% image patches in each layer.  }
\label{fig3}
\centering
\end{figure}

\begin{figure}[h]

\centering
\includegraphics[width=0.45\textwidth]{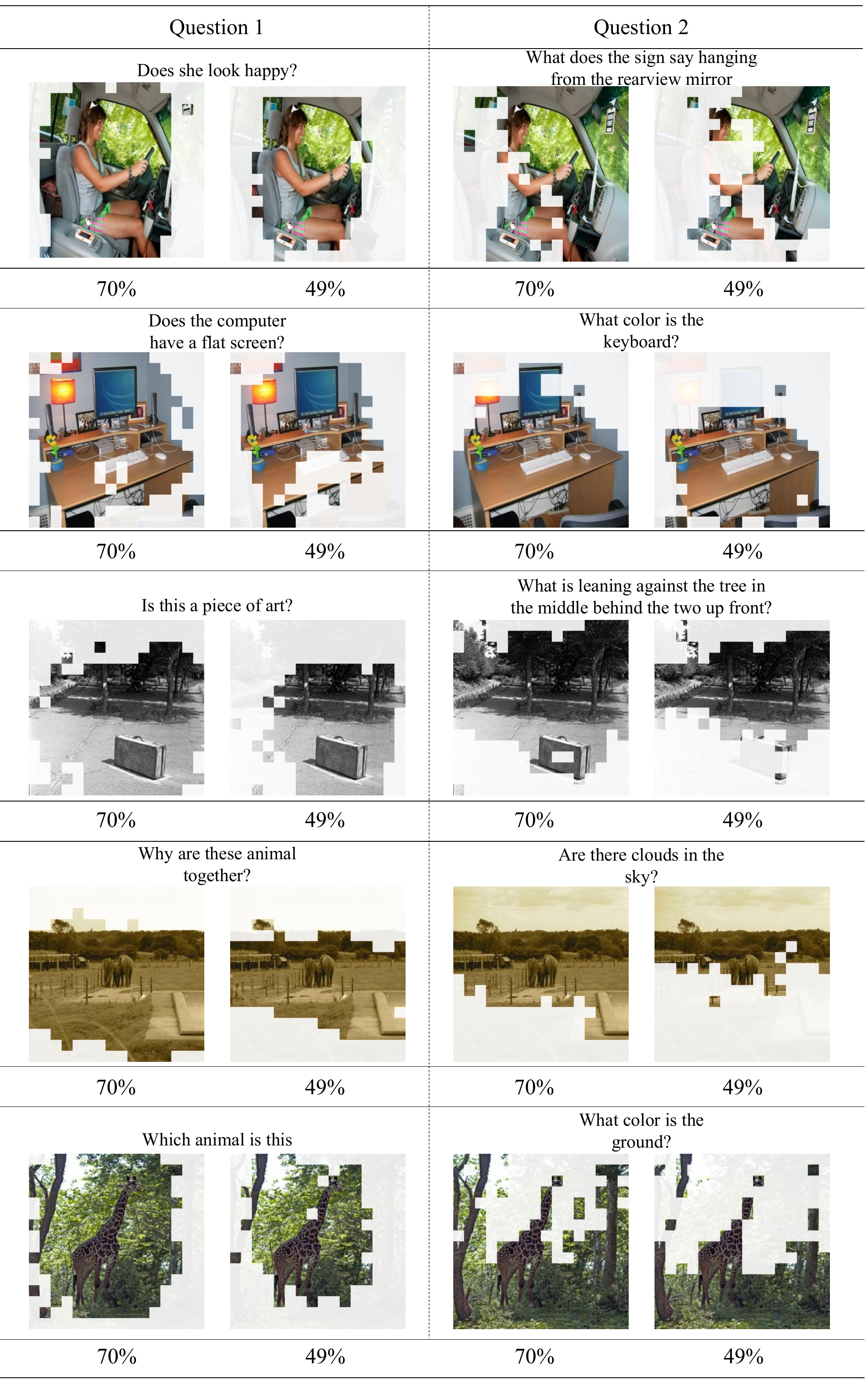}%fig2文件夹下的xbee.esp图片，
\caption{The visualization of the selected text-aware image patches in different selection layers of TRIPS-mPLUG, conditioned on different text queries in the VQA scenario. We set the 5th and 10th layers in the vision backbone as the patch-selection layer, and we keep 70\% image patches in each layer. }
\label{case_vqa}
\centering
\end{figure}
\subsection{Ablation Study}
We also conduct ablation studies to investigate the effects of inattentive image token fusion and Text-aware Dynamic Attention (TD-ATT). In Table \ref{table6}, $w/o$ ITF denotes that the inattentive tokens are directly discarded without fusion. As demonstrated in Table \ref{table6}, fusing inattentive tokens leads to better performance compared to the model without inattentive tokens. Although the improvement is modest, no additional computational overhead is introduced. We further examine the impact of Text-aware Dynamic Attention. Specifically, $w/o$ TD-ATT indicates that we remove the TD-ATT from the patch-selection layer and select the image tokens based on the image [CLS] token. As TRIPS-ViLT is a single-stream model and the [CLS] token aggregates both image and text information, we cannot perform this ablation for it. As demonstrated in Table \ref{table6}, selecting image patch tokens with the image [CLS] token without considering the linguistic context leads to a decline in the model's performance. This result supports our initial motivation that directly removing patch tokens based on the image [CLS] without incorporating text knowledge is unsuitable for VLP models.

\subsection{Visualization}

The proposed TRIPS accelerates VLP through a novel patch selection module that identifies text-consistent image tokens in the vision backbone and preserves the attentive image tokens. To further investigate our model's interpretability, we visualize the process of text-relevant image patch selection during both the pre-training and inference stages.

Figure \ref{fig3} displays the selected text-relevant image patches during the pre-training stage. It is evident that as the network goes deeper, the inattentive tokens are gradually removed or fused, while the text-relevant tokens are selected and preserved.

Figure~\ref{case_vqa} visualizes the selected text-aware image patches in the VQA scenario to demonstrate the effectiveness of the text-relevant image patch selection module during the downstream task inference stage. We input various queries and visualize the text-aware image patches chosen by the text-relevant image patch selection module. As depicted in Figure~\ref{case_vqa}, the selected image patches are highly relevant to the query texts, enabling our model to make accurate predictions.

\section{Conclusion}
We have presented TRIPS, an efficient VLP model with \textbf{T}ext-\textbf{R}elevant \textbf{I}mage \textbf{P}atch \textbf{S}election, which progressively eliminates redundant image tokens with the guidance of text. TRIPS incorporates a novel patch selection module within the vision backbone to select text-consistent image tokens. This module preserves attentive image tokens with text guidance and combines inattentive tokens into a single token by dynamically computing text-dependent visual attention in an end-to-end manner. Experiments demonstrate that our method not only reduces the computational cost of VLP but also enhances the efficiency of cross-modal fusion due to the decreased visual sequences, while maintaining or even improving the performance of downstream image-text tasks.




\bibliography{sn-bibliography}% common bib file





% LNCS Homepage, \url{http://www.springer.com/lncs}. Last accessed 4
% \end{thebibliography}
\clearpage

\begin{appendices}

\section{Comparison Models}
\label{sup:comparison models}
\begin{itemize}

\item \textbf{E2E-VLP}~\citep{Xu2021E2EVLPEV}: proposes the first end-to-end VLP method for both V+L understanding and generation, with a unified Transformer encoder-decoder architecture.

\item \textbf{VinVL}~\citep{2021VinVL}: pre-trains a large-scale object-attribute detection model with much larger amounts of supervised data on four public object detection datasets for extracting better region-based visual features. 

\item \textbf{OSCAR}~\citep{Li2020OscarOA}: proposes to use object tags detected in images as anchor points to ease the learning of cross-modal alignments, where the input to the Transformer is a combination of image, text, and object tags.

\item \textbf{METER}~\citep{dou2021empirical}: systematically investigates how to design and pre-train a fully transformer-based VL model in an end-to-end manner.

\item \textbf{VLMo}~\citep{Wang2021VLMoUV}: presents a unified vision-language pre-trained model that jointly learns a dual encoder and a fusion encoder with a modular Transformer network.

\item \textbf{SimVLM}~\citep{Wang2021SimVLMSV}: different from previous VLP methods that only use limited (4M-10M) image-text pairs for pre-training, it proposes a simple VLP model with a single prefix language modeling objective, which pre-trains on an extremely large aligned cross-modal data of about 1.8B noisy image-text pairs. This is also the latest state-of-the-art method of image captioning.

\item \textbf{ALBEF}~\citep{Li2021AlignBF}: introduces a contrastive loss to align the image and text representations before fusing them through cross-modal attention, which enables more grounded vision and language representation learning.

\item \textbf{UNITER}~\citep{Chen2020UNITERUI}: proposes an improved single-stream VLP method, by designing two new pre-training strategies: 1) it uses conditional masking on pre-training tasks instead of random masking strategy, 2) it designs a new word-region alignment pre-training task via the use of optimal transport to explicitly encourage fine-grained alignment between words and image regions. 

\item \textbf{ALIGN}~\citep{jia2021scaling}: leverages a noisy dataset of over one billion image alt-text pairs, obtained without expensive filtering or post-processing steps in the Conceptual Captions dataset.

% \item \textbf{VLBERT}~\citep{Su2020VLBERTPO}: is a pioneering work to pre-train a single-stream multi-modal Transformer, which jointly trains both the Transformer-based cross-modal fusion and Fast R-CNN image feature extractor in both pre-training and fine-tuning phases. It is widely used as a baseline method for VLP models. 

\item  \textbf{ViLT}~\citep{Kim2021ViLTVT}:  adopts linear projection and word embedding as the visual and textual encoders, and uses the visual transformer as the cross-modal encoder to align and fuse the features of both modalities in an end-to-end manner.

% \item \textbf{XVLM}~\citep{Zeng2021xvlm}: proposes to learn multi-grained alignments which locates visual concepts in the image given the associated texts, and in the meantime align the texts with the visual concepts.
\item \textbf{BLIP}~\citep{li2022blip}: proposes a new VLP framework that transfers flexibly to both vision-language understanding and generation tasks. It effectively utilizes noisy web data by bootstrapping the captions.
\item \textbf{UNICORN}~\citep{DBLP:journals/corr/abs-2111-12085}: proposes a vision-language (VL) model that unifies text generation and bounding box prediction into a single architecture. 

\item \textbf{LXMERT}~\citep{Tan2019LXMERTLC}: is the pioneering work to pre-train a dual-stream multi-modal Transformer, which consists of an object relationship encoder, a language encoder, and a cross-modality encoder. It is widely used as a baseline method for VLP models. 

\item \textbf{ViLBERT}~\citep{Lu2019ViLBERTPT}: proposes the first work that extends the BERT architecture to a multi-modal dual-stream VLP model, which processes both visual and textual inputs in separate streams that interact through co-attentional transformer layers.

\item \textbf{mPLUG} ~\citep{li2022mplug}: is a  vision-language foundation model for both cross-modal understanding and generation and introduces an effective and efficient vision-language architecture with novel cross-modal skip-connections.

\end{itemize}

\section{Downstream Task Details}
\label{sup:downstream task details}

We evaluate TRIPS on the four downstream vision-language tasks. The hyperparameters that we use for finetuning on the downstream tasks are listed in Table \ref{table:finetune-hyper}. Following ~\citep{Li2021AlignBF}, all tasks adopt RandAugment, AdamW optimizer with a weight decay of 0.05 and a cosine learning rate schedule. Next, we introduce the dataset settings in detail.

% \subsection{Ablation Studies of Pre-training Objectives}

\paragraph{VQA.} The VQA task ~\citep{Agrawal2015VQAVQ} requires the model to answer natural language questions given an image. Most methods~\citep{Tan2019LXMERTLC,Wang2021VLMoUV,Li2020OscarOA,Wang2021SimVLMSV} deal with visual question answering tasks as multi-label classification on pre-defined answer sets. This strategy achieves strong performance, but it is not suitable for real-world open scenarios. We conduct an experiment on the VQA2.0 dataset~\citep{goyal2017making}, which contains 83k/41k/81k images for training/validation/test. Following ~\citep{Li2021AlignBF}, we use both training and validation splits for training, and incorporate additional training data from Visual Genome~\citep{Krishna2016VisualGC}. Following \citep{Li2020OscarOA}, we concatenate the question with the object labels and OCR tokens extracted from the image.
% Moreover, we directly select the silent image according to the ROUGE-L scores between the aligned sentences or image caption
\begin{table}
\caption{Finetuning hyperparameters for downstream tasks. $\dagger$ denotes two stages of fine-tuning.}
\setlength\tabcolsep{1pt}
\centering
\footnotesize
\begin{tabular}{l|ccc}
\toprule[1.5pt]
Task  &  LR (ViT-L/BERT$_{base}$) & batch size & epochs  \\
\midrule
VQA & 2e-5/5e-6 & 1024 &  8 \\
Captioning$\dagger$ & 1e-5\&8e-7 & 256& 5 \\
Retrieval & 1e-5/2e-6 & 256& 5 \\
NLVR2 & 5e-5/5e-6 & 256 & 15 \\
Visual Grounding & 2e-5/2e-6 & 512& 120 \\
\bottomrule[1.5pt]
\end{tabular} 

\label{table:finetune-hyper}
\end{table}

\paragraph{Image Captioning.} Image captioning requires generating a descriptive and fluent caption for a given image. We evaluate the performance of TRIPS on two popular datasets: COCO Caption~\citep{Lin2014MicrosoftCC} and NoCaps~\citep{nocaps}. We fine-tune TRIPS-mPLUG on the training set of COCO Caption and test it on the same Karpathy split~\citep{Li2020OscarOA,Wang2021SimVLMSV} as well as the NoCaps validation set. To fine-tune TRIPS-mPLUG on COCO Caption, we follow the approach in~\citep{Li2020OscarOA} and first train the model with the cross-entropy loss for 5 epochs with a learning rate of 1e-5 and a batch size of 256. We then further fine-tune the model with CIDEr optimization~\citep{scst} for an additional 5 epochs with a smaller learning rate of 8e-7. We use the best checkpoint on COCO Caption to predict on the NoCaps validation set.
During inference, we use beam search with a beam size of 10 and set the maximum generation length to 20.

% \paragraph{Image-Text Retrieval.} We conduct experiments for both image-to-text retrieval (TR) and text-to-image retrieval (IR) on COCO ~\citep{Lin2014MicrosoftCC} and Flickr30K ~\citep{Plummer2015Flickr30kEC} datasets. We adopt the widely-used Karpathy split ~\citep{karpathy2015deep} for both COCO and Flickr30K. COCO contains 113k/5k/5k images for train/validation/test, and Flickr30K contains 29k/1k/1k images for train/validation/test. Following ~\citep{Li2021AlignBF, li2022blip}, we jointly optimize the ITC loss and the ITM loss during fine-tuning. During inference, we first select top-k candidates by computing the dot-product similarity between the image and text encoder features (When extracting the image encoder feaure, for efficiency of coarse-grained ranking, we replace the SPD with a simple strategy which we directly detect the patch in a transformer layer based on the self-attention weights of image [CLS] token to other patch tokens ),and then rerank the selected candidates based on their ITM scores (In the fine-grained reranking stage, for the same image, we re-extracting multiple image encoder features based on SPD with the guidence of multiple text candidates.). We set $k = 256$ for COCO and $k = 128$ for Flickr30K.
\paragraph{Image-Text Retrieval.} We conducted experiments on both image-to-text retrieval (TR) and text-to-image retrieval (IR) using the COCO~\citep{Lin2014MicrosoftCC} and Flickr30K~\citep{Plummer2015Flickr30kEC} datasets and used the widely-used Karpathy split~\citep{karpathy2015deep} for both. COCO contains 113k/5k/5k images for train/validation/test, while Flickr30K contains 29k/1k/1k images for train/validation/test. During fine-tuning, for TRIPS-ALBEF and TRIPS-mPLUG we jointly optimized the ITC loss and the ITM loss following the approach in~\citep{Li2021AlignBF, li2022blip}. During inference, we first selected the top-k candidates by computing the dot-product similarity between the image and text encoder features (We set $k=256$ for COCO and $k=128$ for Flickr30K). For the efficiency of coarse-grained ranking, we directly selected the patch based on the attention weights of the image [CLS] token to other patch tokens. 

\paragraph{Visual Grounding.} The task of visual grounding involves localizing the referred object in an image given a plain text query. Instead of directly regressing bounding boxes, our approach concatenates visual features with textual features, which are then fed into the multi-modal decoder to predict the object's coordinates. We evaluate our method on the referring expression grounding dataset: RefCOCO+\citep{yu2016modeling}. The RefCOCO+ dataset contains 19K images and 141K queries.

\paragraph{NLVR2.} The NLVR2~\citep{Suhr2019ACF} task requires the model to predict whether a sentence. We conduct experiments following the original train/val/test split in ~\citep{Suhr2019ACF}. Following \citep{li2022blip}, we use two cross-attention layers to process the two input images, and their outputs are merged and fed to the FFN. An MLP classifier is then applied to the output embedding of the language [CLS] token.




\end{appendices}

\end{document}